\documentclass[sigconf]{acmart}
\usepackage{multirow}
\usepackage{colortbl}
\usepackage{subfig}
\usepackage{tabularx}
\usepackage{subfig}
\usepackage{pifont}
\usepackage{bbding}
\AtBeginDocument{%
  }

\copyrightyear{2024}
\acmYear{2024}
\setcopyright{rightsretained}
\acmConference[CIKM '24]{Proceedings of the 33rd ACM International Conference on Information and Knowledge Management}{October 21--25, 2024}{Boise, ID, USA}
\acmBooktitle{Proceedings of the 33rd ACM International Conference on Information and Knowledge Management (CIKM '24), October 21--25, 2024, Boise, ID, USA}
\acmDOI{10.1145/3627673.3679967}
\acmISBN{979-8-4007-0436-9/24/10}

\begin{document}

\title{MSG-Chart: Multimodal Scene Graph for ChartQA}
\author{Yue Dai}
\email{yue.dai@research.uwa.edu.au}
\orcid{0000-0003-1160-7927}
\affiliation{%
  \institution{The University of Western Australia}
  \streetaddress{P.O. Box 1212}
  \city{Perth}
  \state{Western Australia}
  \country{Australia}
  \postcode{43017-6221}
}

\author{Soyeon Caren Han}
\authornote{Corresponding author.}
\affiliation{%
  \institution{The University of Melbourne}
  \streetaddress{1 Th{\o}rv{\"a}ld Circle}
  \city{Melbourne}
  \country{Australia}}
\email{caren.han@unimelb.edu.au}

\author{Wei Liu}
\affiliation{%
  \institution{The University of Western Australia}
  \streetaddress{1 Th{\o}rv{\"a}ld Circle}
  \city{Perth}
  \country{Australia}}
\email{wei.liu@uwa.edu.au}

\renewcommand{\shortauthors}{Yue, et al.}

\begin{abstract}
Automatic Chart Question Answering (ChartQA) is challenging due to the complex distribution of chart elements with patterns of the underlying data not explicitly displayed in charts.
To address this challenge, we design a joint multimodal scene graph for charts to explicitly represent the relationships between chart elements and their patterns. 
Our proposed multimodal scene graph includes a visual graph and a textual graph to jointly capture the structural and semantical knowledge from the chart. 
This graph module can be easily integrated with different vision transformers as inductive bias. 
Our experiments demonstrate that incorporating the proposed graph module enhances the understanding of charts' elements' structure and semantics, thereby improving performance on publicly available benchmarks, ChartQA and OpenCQA.\footnote{Code available at \url{https://github.com/adlnlp/MSG-Chart}}

\end{abstract}




\maketitle

\section{Introduction}
\begin{figure}
    \centering
    \includegraphics[width=\columnwidth]{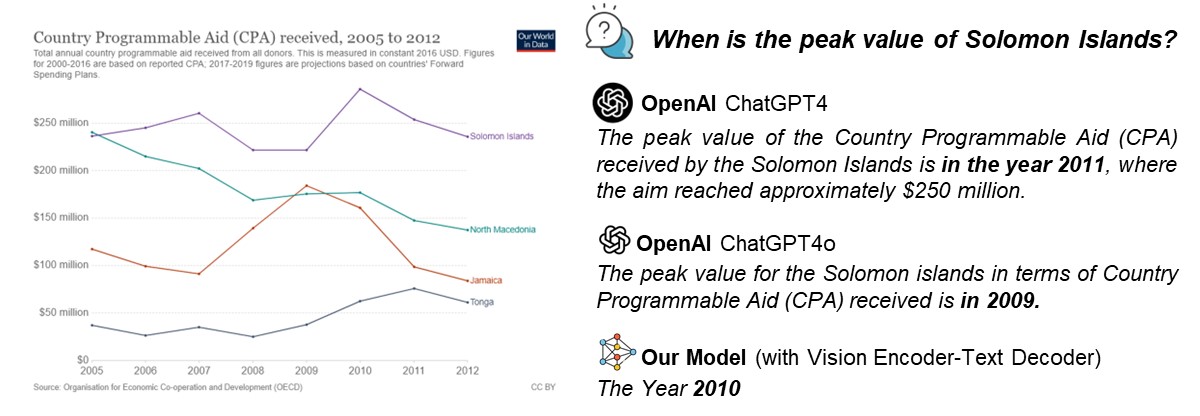}
    \caption{Cutting-Edge LLMs and Our MSG-Chart}
    \label{fig:llmhallu}
    \vspace{-0.5cm}
\end{figure}

Chart Question Answering (ChartQA) involves answering questions presented in natural languages based on information from a chart. As a subset of Visual Question Answering (VQA), ChartQA presents unique challenges distinct from those encountered with natural images. Firstly, a chart contains numerous elements and various element types within a single image. Although the distribution of these elements can be complex, charts are generally highly structured. For instance, y axis labels are normally located at the left side of the image while x axis labels are at the bottom. Secondly, charts often include extensive text and numerical data, understanding these features is crucial for accurate question answering. While recognizing the underlying text of an object is enough for data extraction tasks, more complex reasoning requires understanding the relationships between elements. For example, in Fig. \ref{fig:llmhallu}, the model must recognise that `Solomon Islands' is not just a label but specifically the label for the purple line. This highlights the importance of comprehending the structure and the relationships within charts.

Although various models have been proposed to address ChartQ-A, they have not effectively tackled the challenges above. \cite{DBLP:conf/acl/0001EPKPLJCCA23} introduced DePlot for chart-to-table conversion. The converted textual data table from the chart is then fed to Flan-PaLM \cite{chung2024scaling} for question answering. Despite the strong mathematical reasoning capabilities of Large Language Models (LLMs), this approach struggles with questions related to visual attributes, such as colour, due to the lack of visual information understanding in textual tables. \citet{DBLP:conf/acl/MasryLTJH22} and \citet{DBLP:conf/acl/TangBS23} employ VL-T5 \cite{DBLP:conf/icml/ChoLTB21} for ChartQA and Chart Captioning, utilizing Mask R-CNN \cite{DBLP:conf/iccv/HeGDG17} or Faster R-CNN \cite{DBLP:conf/nips/RenHGS15} to extract Region of Interest (ROI) features of objects, combined with textual data tables or scene graphs as input. However, these methods lack explicit spatial relationships between ROI features, and the connections between visual features and their underlying value remain unclear. More recently, pre-trained multimodal language models \cite{DBLP:conf/acl/0001PKPLJACE23, DBLP:conf/emnlp/MasryKLHJ23} have been proposed.  Their patch-based image inputs can result in chart elements being split across different patches, losing object-wise information and making it challenging for the model to capture the full pattern of an intact object. This limitation is also present in the latest LLMs, such as GPT-4 and GPT-4o, as demonstrated in Fig. \ref{fig:llmhallu}. While these models generate detailed answers, they struggle to identify peak values and correctly estimate values not explicitly listed on the y-axis labels.
\begin{figure*}[tbh]%
    \centering
    \subfloat[\centering Vision Encoder-Text Decoder (Backbone: UniChart \cite{DBLP:conf/emnlp/MasryKLHJ23})\label{fig:unichart_archs}]{{\includegraphics[width=0.9\columnwidth]{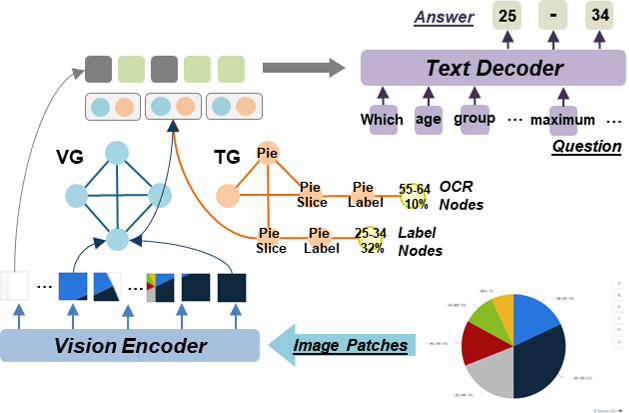} }}%
    \qquad
    \hspace{1cm}
    \subfloat[\centering Multimodal Encoder-Text Decoder (Backbone: VL-T5\cite{DBLP:conf/icml/ChoLTB21})\label{fig:vlt5_archs}]{{\includegraphics[width=0.88\columnwidth]{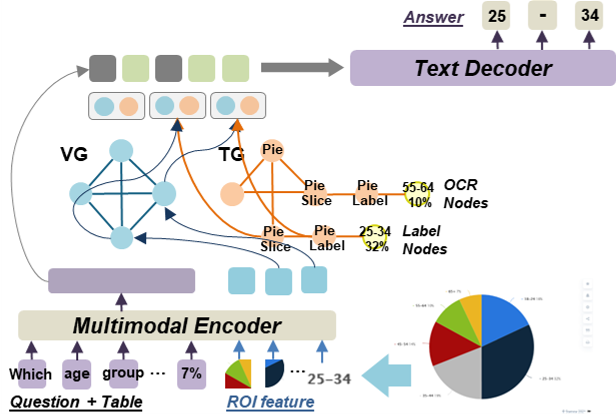} }}%
    \caption{Two Graph Integration Architectures (VG: Visual Graph, TG: Textual Graph)}%
    \label{fig:archs}%
\end{figure*}
Instead of implicitly learning structural and semantic information from charts using pre-trained models, we explicitly use graphs to capture the structural and semantic information from charts. To do so, we propose the multimodal scene graph for chart understanding with visual and textual graphs. The visual scene graph captures the spatial relation in terms of visual features, and the textual scene graph captures the semantic knowledge using textual features. Inspired by \cite{han2020victr,DBLP:conf/aaai/LiHCQ0HHDSL23}, we integrated the graph representation as the inductive bias to the backbone model. Our contributions are: 1) We propose new multimodal scene graphs for chart understanding to represent the structural and semantic relation. 2) The proposed joint multimodal scene graph can be adapted to diverse chart understanding backbones. 3) Our multimodal scene graph can enhance the performance of the diverse ChartQA backbone models.

\section{Methodology}
\subsection{Graph Construction}
Figure \ref{fig:archs} presents the proposed graph-based model on two backbone frameworks: 1) Vision Encoder-Text Decoder (UniChart \cite{DBLP:conf/emnlp/MasryKLHJ23}) and 2) Multimodal Encoder-Text Decoder (VL-T5 \cite{DBLP:conf/icml/ChoLTB21})
The proposed model includes two undirected graphs: visual and textual graphs. The visual graph is initialised using the hidden states from the encoder, capturing structural information related to visual aspects. The textual graph uses chart elements' labels and OCR features, capturing semantic knowledge. 

\textbf{Visual Graph} ($G_V$) is a fully connected graph designed to capture the spatial relations of objects from a visual aspect. Each node represents an object detected by the Mask R-CNN \cite{DBLP:conf/iccv/HeGDG17} fine-tuned by \citet{DBLP:conf/acl/MasryLTJH22}.
To maintain the spatial relationships between objects, inspired by the basis decomposition method introduced in R-GCN \cite{DBLP:conf/esws/SchlichtkrullKB18}, we assign each edge $e$ a coefficient $a_e = \exp(-d)$, where $d$ represents the smallest Euclidean distance between the bounding boxes of two objects. This coefficient prioritises closer neighbours of a node by assigning them larger weights. 
With each object's bounding box, we can determine the alignment between image patches and corresponding objects. After that, each node is initialised 
as $V_o = \frac{1}{|P_o|} \sum_{p\in P_o}^{P_o} H_{e_p}$, where $P_o$ represents the set of patches that the object $o$ is located, and $H_e$ is the last hidden states from the encoder of the backbone model. 
For the vision encoder backbone in Figure \ref{fig:unichart_archs}, the blue visual node at the bottom represents the dark blue pie slice in the chart image. It is initialised by the mean of hidden states from the corresponding patches. For the multimodal encoder backbone in Figure \ref{fig:vlt5_archs}, each ROI represents an intact object so the same visual node is directly initialised by the hidden states representing the dark blue pie slice.

\textbf{Textual graph} ($G_t$) is constructed with textual features based on chart elements. The graph has two types of nodes: label nodes and OCR nodes. Following scene graphs \cite{wu-etal-2023-scene, yoon2023unbiased, shukla2023scene}, each label node represents an object's label predicted by Mask R-CNN. Label nodes are connected based on the chart semantics. The x/y axis title connects to the x/y axis labels representing x/y axis values. X/Y axis labels connect to lines, bars, or dot lines if their bounding box values overlap on longitude/latitude, indicating value ranges of the latter. Legend labels connect to the closest legend marker to show the former, which is the label of chart elements with the same colour as the legend marker. Pie charts connect to all slices, representing the whole-part relationship, while pie labels connect the nearest pie slice to show the former is the latter's label.
OCR nodes represent the OCR text extracted from non-shape objects such as the title and x-axis label. Following \citet{DBLP:conf/wacv/MethaniGKK20}, we utilise the open-sourced Optical Character Recognition (OCR) model Tesseract\footnote{\url{https://github.com/tesseract-ocr/tesseract}} to extract texts from the cropped image of each detected object. These OCR nodes are linked to their corresponding label nodes. This allows the textual graph to capture not only the OCR texts from charts but also the relationships between the underlying data and objects. All nodes in the semantic graph are initialised with the Mean of BERT \cite{DBLP:conf/naacl/DevlinCLT19} embeddings of the text they represent.
Given the graphs $G_v$, $G_t$, visual features $V$, and textual features $T$ that have been projected to the same space as $V$, we obtain graph representations $H_v = G_v(V)$ and $ H_t = G_t(T) $ using GCN \cite{DBLP:conf/iclr/KipfW17}. Afterward, we concatenate the graph representations of the same object to combine the structural and semantic information, for example in Fig. \ref{fig:archs}, the nodes from visual and textual graph representing the dark blue pie slice and concatenated together. We then utilise the multilayer perceptron with the ReLU activation function to project the concatenation into the space of the hidden states of the backbone model as $H_G$.

\subsection{Graph Integration and ChartQA}
Inspired by \cite{DBLP:conf/aaai/LiHCQ0HHDSL23}, we inject the graph representation as the inductive bias to the backbone model. We ensure that our proposed multimodal scene graph module functions can be a general plug-in component without substantially modifying the backbone model. Hence, we integrate the graph module between the encoder and text decoder. 
We tested our methods on the following two types of backbone models, Vision Encoder-Text Decoder (UniChart\cite{DBLP:conf/emnlp/MasryKLHJ23}) and Multimodal Encoder-Text Decoder (VL-T5\cite{DBLP:conf/icml/ChoLTB21}).

\textbf{Vision Encoder-Text Decoder}: UniChart\cite{DBLP:conf/emnlp/MasryKLHJ23} is a state-of-the-art open-source Vision Encoder-Text Decoder model for chart comprehension. UniChart utilises the Donut \cite{DBLP:conf/eccv/KimHYNPYHYHP22} encoder as the image encoder and the BART \cite{DBLP:conf/acl/LewisLGGMLSZ20} decoder as the text decoder. The models with this architecture take sequential image patches as encoder input and question tokens as input for the decoder. To inject entity-related structural and semantical information into the model, we fuse the same node representation to the patches where the object is located. For example, in Figure \ref{fig:unichart_archs}, the updated node representation of the dark blue slice from the graph module should be fused with the hidden states where the patches represent the same object. Specifically, we create a bias representation $ H_b $ based on graph representation $ H_G $. For an index $ I $ and an object set $ O_i $, where the bounding box of every object $o\in O_i$ intersects with patch $ p_i $, the feature of $ H_{b_i} $ is assigned by the mean of $ H_{G_{O_i}} $. Formally,

\begin{equation}
        H_{b_i}=\left\{\begin{matrix}
\frac{1}{|O_i|}\sum _{i=1}^{O_i}H_{G_i} & \text{if}\quad |O_i|>0 \\ 
0 & \text{if}\quad |O_i|=0
\end{matrix}\right.
\label{eq:hb}
\end{equation}

\textbf{Multimodal Encoder-Text Decoder}: The second backbone model is VL-T5, a widely used Multimodal Enocder-Text Decoder model employed in prior research for ChartQA and Chart Summarisation tasks \cite{DBLP:conf/acl/MasryLTJH22, DBLP:conf/acl/TangBS23, DBLP:conf/emnlp/KantharajDLTHJ22}. VL-T5 takes both textual tokens and ROI features as input. Since each ROI feature represents one object, we fuse the node representations with their corresponding hidden states. In the case of Figure \ref{fig:vlt5_archs}, the updated node representation of the dark blue pie slice will be fused with the second visual token (blue rectangle) from the encoder.  The new representation $H_b$ is a special case of Eq. \ref{eq:hb} where $|O_i| = 1$. Since VL-T5 is pre-trained with 36 object regions, we select the top 36 objects detected by Mask R-CNN ranked by confidence score. If fewer than 36 objects are detected, we pad the input with zeros to a fixed length of 36. The original hidden states from the encoder $ H_e $ are updated by adding the bias $H_b$, resulting in $ \widetilde{H}_e = H_e + H_b $. The new hidden states are then passed to the decoder for answer generation.

Given an image $ I $, texts $ x $, and label $ y $, we jointly train the parameters of the backbone model $ \theta_m $, visual and textual graphs $ \theta_{v}, \theta_{t} $ by minimizing the negative log-likelihood:
\begin{equation}
   L_{\theta_m, \theta_{v}, \theta_{t}}=-\sum ^{\left | y \right |}_{j=1}\log P_{\theta_m, \theta_{v}, \theta_{t}}(y_j|y_{<j},x,I)
\end{equation}

\section{Experiment}
% \subsection{Dataset and Evaluation Metrics}
We tested our propose model on ChartQA and OpenCQA: \textbf{ChartQA} is currently the most challenging Chart Question Answering dataset. The questions in this dataset are either generated by a fine-tuned T5 \cite{DBLP:journals/jmlr/RaffelSRLNMZLL20} model or through human annotation, resulting in two sets: augmentation and human. The questions in the human set are more difficult as they emphasise logical and visual reasoning. Following \cite{DBLP:conf/acl/0001PKPLJACE23, DBLP:conf/acl/MasryLTJH22, DBLP:conf/wacv/MethaniGKK20}, we use relaxed accuracy, which requires an exact match for textual answers and allows a 5\% tolerance for numerical answers. 
\textbf{OpenCQA} \cite{DBLP:conf/emnlp/KantharajDLTHJ22} is an open-ended ChartQA dataset. Unlike previous ChartQA datasets, where the answers are mainly words or phrases, the answers in OpenCQA are explanatory texts, with an average answer length of 56 tokens. We use BLEU4 \cite{DBLP:conf/wmt/Post18} following \cite{DBLP:conf/emnlp/KantharajDLTHJ22, DBLP:conf/emnlp/MasryKLHJ23}. We scale to the range of 0 to 100, consistent with previous research.

% \subsection{Implementation Details}
We have different text input settings for different models and datasets.
\textbf{ChartQA} For VL-T5, we use the flattened ground truth table and question tokens as text inputs. This is to ensure fair comparison with the performance in \cite{DBLP:conf/acl/MasryLTJH22}, because the model they used to generate table from chart automatically is not available. For UniChart, we only use questions as text input.
\textbf{OpenCQA} The original work in \cite{DBLP:conf/emnlp/KantharajDLTHJ22} has three settings. We use the setting where chart images, questions and OCR texts are used as input. We use this setting for VL-T5 as it's the one used in comparison with UniChart paper. For UniChart, the text input is the question.

All experiments were conducted using a single Nvidia A100 (40G) GPU, with the random seed set to 42 for reproducibility. We use AdamW as the optimiser. The settings for VL-T5 on both datasets follow the original papers \cite{DBLP:conf/acl/MasryLTJH22} and \cite{DBLP:conf/emnlp/KantharajDLTHJ22}, with the exception that the batch size for ChartQA is 24 and for OpenCQA is 12 due to GPU memory limitations. The settings for UniChart follow \cite{DBLP:conf/emnlp/MasryKLHJ23}, with the model trained using mixed precision consistent with the provided code. The batch sizes for both datasets are set to 8 due to GPU memory constraints. Additionally, the number of training epochs for OpenCQA is set to 20, different from the original paper, because we observed overfitting with 20 epochs. The GCNs for both the visual and textual graphs consist of 2 layers, and the hidden state dimensions matched the backbone models: 1024 for UniChart and 768 for VL-T5. The dropout rate is set to 0.2.

\begin{table}[tbh]
\resizebox{\columnwidth}{!}{\begin{tabular}{lllrrrr}
\toprule
                               &     &     & \multicolumn{3}{c}{\textbf{ChartQA}} & \multicolumn{1}{c}{\textbf{OpenCQA}} \\
 &
  \textbf{GT} &
  \textbf{Graph} &
  \multicolumn{1}{r}{\textbf{aug.}} &
  \multicolumn{1}{r}{\textbf{human}} &
  \multicolumn{1}{r}{\textbf{avg.}} &
  \multicolumn{1}{r}{\textbf{BLEU}} \\ \midrule
Pix2Struct \cite{DBLP:conf/icml/LeeJTH0EKSCT23}  &  $\times$  &  $\times$  & 81.60   & 30.50   & 56.00   & -                           \\
Matcha\footnotemark[3] \cite{DBLP:conf/acl/0001PKPLJACE23} &  $\times$  &  $\times$  & 90.20   & 38.20   & 64.20   & -                           \\
Matcha\footnotemark[4]         &  $\times$  &  $\times$  & 86.64   & 36.96   & 61.80   & -                           \\
Matcha\footnotemark[5]         &  $\times$  &  $\times$  & 81.28   & 28.16   & 54.72   & -                           \\ \midrule
UniChart\footnotemark[3] \cite{DBLP:conf/emnlp/MasryKLHJ23}      &  $\times$  &  $\times$  & 88.56   & 43.92   & 66.24   & 14.88                       \\
UniChart\footnotemark[4]       &  $\times$  &  $\times$  & 82.32   & 34.48   & 58.40   & 8.76                        \\
UniChart\footnotemark[5]       &  $\times$  &  $\times$  & 82.00   & 30.80   & 56.40   & 10.86                       \\
\textbf{UniChart (Ours)}       & $\times$  & \checkmark & \textbf{85.36} & \textbf{37.44} & \textbf{61.4}  & \textbf{11.97} \\
\midrule
VL-T5 \cite{DBLP:conf/acl/MasryLTJH22, DBLP:conf/emnlp/KantharajDLTHJ22} & \checkmark &  $\times$  & -       & -       & 59.12   & 14.73                       \\ 
\textbf{VL-T5 (Ours)}          & \checkmark & \checkmark & \textbf{92.4} & \textbf{38} & \textbf{64.8} & \textbf{17.14} \\ \bottomrule
\end{tabular}}
\caption{Overall Performance on ChartQA (GT: Golden Table)}
\label{tab:res}
\end{table}

\footnotetext[3]{Result from original paper}
\footnotetext[4]{Result from checkpoint in huggingface}
\footnotetext[5]{Result from model fine-tuned by ourselves}

\vspace{-0.5cm}
\section{Results}
\subsection{Overall Performance}
We present the results of our models compared with previous baselines in Table \ref{tab:res}. Note that the performance of UniChart differs from the original paper's reported results. The publicly available code or the given huggingface checkpoint\footnote{\url{https://huggingface.co/ahmed-masry/unichart-base-960}} provided by \cite{DBLP:conf/emnlp/MasryKLHJ23} are unable to reproduce the same performance, which yields an accuracy of 58.4 instead of the reported 66.24. This discrepancy may be due to differences in the environment, such as GPU configuration and random seed. For a fair comparison, we fine-tuned the model in our environment, which resulted in a 56.4 overall accuracy. Similarly, for UniChart on OpenCQA, the result from the provided checkpoint is 8.76 rather than the reported 14.88 from the paper, and the result fine-tuned by ourselves is 10.86. Under the same training environment, our proposed graph module enhances the performance of both UniChart and VL-T5 on both datasets.  Specifically, UniChart demonstrates at least a 3\% improvement in accuracy on ChartQA and a 1.1-point increase in BLEU score on OpenCQA. For VL-T5, the accuracy on ChartQA increases by 5.68\%, and the BLEU score on OpenCQA rises by 2.41 points.   
The produced result trend shows that integrating our components with any state-of-the-art chartQA framework is feasible. It also proves the superiority of our proposed graph model, which consistently improves overall performance in diverse types of chartQA tasks. 

\subsection{Ablation Studies}

\begin{table}[bh]
\small
\begin{tabular}{lllrrrr} \toprule
\textbf{} & \textbf{} & \textbf{} & \multicolumn{3}{c}{\textbf{ChartQA}} & \multicolumn{1}{c}{\textbf{OpenCQA}}\\
\textbf{} & \textbf{VG} & \textbf{TG} & \multicolumn{1}{r}{\textbf{aug.}} & \multicolumn{1}{r}{\textbf{human}} & \multicolumn{1}{r}{\textbf{avg.}} & \multicolumn{1}{r}{\textbf{BLEU}} \\ \midrule
\multirow{4}{*}{UniChart} & \checkmark & \checkmark & \textbf{85.36} & \textbf{37.44} & \textbf{61.4} & \textbf{11.97}\\
 & \checkmark &  $\times$ & \underline{84.88} & \underline{35.84} & \underline{60.36} & 11.42\\
 &  $\times$ & \checkmark & 82.8 & 34.56 & 58.68 & \underline{11.76}\\ 
 &  $\times$ &  $\times$ & 82.32 & 34.48 & 58.40 & 8.76\\ \midrule
\multirow{4}{*}{VL-T5} & \checkmark & \checkmark & \textbf{92.4} & \textbf{38} & \textbf{64.8} & \underline{17.14}\\
 & \checkmark &  $\times$ & \underline{91.12} & 36.08 & 63.6 & 16.99\\
 &  $\times$ & \checkmark & 90.56 & \underline{37.36} & \underline{63.96} & \textbf{18.14}\\
 &  $\times$ &  $\times$ & - & - & 59.12 & 14.73\\ \bottomrule
\end{tabular}
\caption{Performance comparison w.r.t. different graph setting (VG: Visual Graph, TG: Textual Graph)}
\label{tab:pgsg}
\end{table}
\vspace{-0.5cm}
\textbf{Effect of the VG and TG:} We conducted ablation studies with different graph settings. As shown in Table \ref{tab:pgsg}, performances are consistently better when both graphs are employed. For UniChart and VL-T5, omitting the visual graph leads to the most significant performance drop on the augmented test set of ChartQA, with decreases of 2.56\% for UniChart and 1.84\% for VL-T5. This aligns with the findings of \cite{DBLP:conf/acl/0001PKPLJACE23} that the augmented set contains more extractive questions, while the human set includes more complex reasoning questions. This demonstrates that our visual graph enhances the models' structural understanding of charts. When only the visual graph is used, the performance on the human set increases by 1.36\% for UniChart. However, when both the visual and textual graphs are utilised, the performance on the human set improves by 2.96\%, indicating that the textual graph significantly enhances the semantic understanding of charts. However, since complex math reasoning questions also require sophisticated structural understanding, using textual graphs alone doesn't necessarily result in the second-best performance (results with underline) on the human set, as seen with VL-T5. Nonetheless, when both graphs are used, the performance on the human set is consistently the best for both models.

\begin{table}[]
\small
\begin{tabular}{llrrrr} \toprule
\multicolumn{2}{l}{\textbf{}} & \multicolumn{3}{c}{\textbf{ChartQA}} & \multicolumn{1}{c}{\textbf{OpenCQA}}\\
\textbf{} & \textbf{TG Relation} & \multicolumn{1}{r}{\textbf{aug.}} & \multicolumn{1}{r}{\textbf{human}} & \multicolumn{1}{r}{\textbf{avg.}} & \multicolumn{1}{r}{\textbf{BLEU}}\\ \midrule
\multirow{2}{*}{UniChart} & designed rules & \textbf{85.36} & \textbf{37.44} & \textbf{61.4} & \textbf{11.97}\\
 & fully connected & 84.24 & 33.44 & 58.84 & 11.40\\ \midrule
\multirow{2}{*}{VL-T5} & designed rules & \textbf{92.4} & \textbf{38} & \textbf{64.8} & 17.14\\
 & fully connected & 91.68 & 34.8 & 63.24 & \textbf{17.92}\\ \bottomrule
\end{tabular}%
\caption{Performance comparison w.r.t. designed semantic relations and fully connected graph (TG: Textual Graph)}
\label{tab:relationsg}
\end{table}

\textbf{Effect of the proposed relation-based graphs:} To evaluate whether the proposed relation in graphs can effectively capture semantic information in charts, we replaced the proposed edges in textual graphs with fully connected edges. The results in Table \ref{tab:relationsg} reveal that fully connected graphs significantly decrease performance on the human set of ChartQA, with a 4\% drop for UniChart and a 3.2\% drop for VL-T5. This difference in the human test set indicates that a carefully designed scene graph enhances the backbone model's ability to understand the semantics of charts. Although OpenCQA does not have significant gap between those with diverse edge types, it is expected as OpenCQA includes more explanatory and descriptive text to answer questions. 

% \usepackage{multirow}
% \usepackage{graphicx}
\begin{table}[]
\small
\begin{tabular}{lllrrrr}
\toprule
\textbf{} & \textbf{} & \textbf{} & \multicolumn{3}{c}{\textbf{ChartQA}} & \multicolumn{1}{c}{\textbf{OpenCQA}}\\
\textbf{} & \textbf{G} & \textbf{Text} & \multicolumn{1}{r}{\textbf{aug.}} & \multicolumn{1}{r}{\textbf{human}} & \multicolumn{1}{r}{\textbf{avg.}} & \multicolumn{1}{r}{\textbf{BLEU}}\\ \midrule
\multirow{4}{*}{UniChart} & \checkmark & label+OCR & \textbf{85.36} & \textbf{37.44} & \textbf{61.4} & \textbf{11.97}\\
 & $\times$ & $\times$ & 82.32 & 34.48 & 58.4 & 8.76\\
 & $\times$ & label & 83.52 & \underline{35.04} & \underline{59.28} & \underline{11.46}\\
 & $\times$ & OCR & \underline{84} & 33.68 & 58.84 & 11.15\\ \midrule
\multirow{4}{*}{VL-T5} & \checkmark & label+OCR & \textbf{92.4} & \textbf{38} & \textbf{64.8} & \underline{17.14}\\
 & $\times$ & $\times$ & - & - & 59.12 & 14.73\\
 & $\times$ & label & \underline{92.32} & \underline{36.08} & \underline{64.2} & 15.05\\
 & $\times$ & OCR & 90.56 & 36.08 & 63.32 & \textbf{17.78}\\ \bottomrule
\end{tabular}%
\caption{Performance comparison with the presence of graph and textual feature only (G: Graph)}
\label{tab:tfg}
\vspace{-0.5cm}
\end{table}

\noindent\textbf{Effect of textual features} We conducted an experiment where the graphs were removed, and the textual features were directly fused into the model. We fused either the object's label or the OCR textual feature with the visual features from the encoder. The results are shown in Table \ref{tab:tfg}. The performances with the presence of the graphs are better compared to solely injecting textual features. Depending on the different textual features, the improvement gained from using graphs is at least 2.12\% for UniChart and 0.6\% for VL-T5 on ChartQA. Notably, fusing the label feature yields higher accuracy than the OCR feature, likely due to the noisy text introduced by the OCR error during text detection.

\section{Conclusion}
This research proposes a novel multimodal scene graph, including a visual graph and a textual graph, to capture the structure and semantic information from charts. The graph module can be easily integrated into different types of chartQA frameworks. Through experiments, we show that the graph module can improve the understanding of charts and consequently perform better on ChartQA tasks. We hope that this multimodal chart-based scene graph can be a useful stepstone to improve the performance of chart understanding and retrieval.

\bibliographystyle{ACM-Reference-Format}
\bibliography{sample-base}

% \appendix

% \section{Appendix}


\end{document}